\documentclass[11pt,a4paper]{article}
\usepackage[hyperref]{emnlp2020}
\usepackage{times}
\usepackage{latexsym}
\usepackage{multirow}
\usepackage{booktabs}
\usepackage{amsmath}
\usepackage{graphicx}
\usepackage{xcolor}

\usepackage{microtype}

\aclfinalcopy 

\title{{\sc OpineSum}: Entailment-based self-training for abstractive opinion summarization \\
{\small }}

\author{
  Annie Louis \\
  Google Research \\
  \texttt{\small annielouis@google.com} \And
  Joshua Maynez \\
  Google Research \\
  \texttt{\small joshuahm@google.com}
}

\date{}

\begin{document}
\maketitle
\begin{abstract}

A typical product or place often has hundreds of reviews, and 
summarization of these texts is an important and challenging problem. 
Recent progress on abstractive summarization in domains such as news
has been driven by supervised systems trained on hundreds of thousands of 
news articles paired with human-written summaries. However for opinion texts, such large scale
datasets are rarely available. Unsupervised methods, self-training, and 
few-shot learning approaches bridge that gap. In this work, we 
present a novel self-training approach, {\sc OpineSum} for \emph{abstractive opinion
summarization}. The summaries in this approach are built using a novel application 
of textual entailment and capture the consensus of opinions across the various reviews for an item. This method can be
used to obtain silver-standard summaries on a large scale and train both 
unsupervised and few-shot abstractive summarization systems. {\sc OpineSum} achieves state-of-the-art performance in both settings. 

\end{abstract}

\section{Introduction}
Understanding the essence of multiple reviews or opinions is a frequent
problem today. High quality opinion summaries can improve search,
product comparison, recommender systems, and even dialog assistants. In this domain, abstractive summarization is
particularly promising for fluently comparing and contrasting opinions from source reviews.
However, while language models trained on huge numbers of source-summary pairs have driven summarization performance 
in some domains, is it harder to find such pairs on the web for opinions, and it is difficult to present tens or
hundreds of reviews to human annotators and train them to write an informative summary. This paper presents a new self-training approach that automatically identifies and leverages \emph{common opinions} across reviews, for example as in Table \ref{tab:introexample}.

\begin{table}[t]
\begin{tabular}{p{0.2cm}|p{6.6cm}}
{\footnotesize R1} & {\footnotesize...very large and clean with a nice size kitchen. {\color{blue}The hotel is located right across the street from balboa park} and within walking distance of a rite aid drugstore..} \\ \hline
{\footnotesize R2} & {\footnotesize...If you insist on staying here, reserve a refurbished room and get that promise in writing! The location was great for tourists, {\color{blue} right across from balboa park}. You could walk to the zoo (about 1/4 mi)...}\\\hline
{\footnotesize R3} & {\footnotesize...I decided to stay at the park manor suites hotel since it seemed to be close to san diego zoo. {\color{blue} The hotel is conveniently located in front of balboa park}, walking distance to san diego zoo,...}\\\hline
{\footnotesize R4} & {\footnotesize...The staff are both pleasant and professional. {\color{blue} Hotel is across from balboa park on sixth ave}. This is the park west area, and features a diverse array of restaurants...}\\\hline
{\footnotesize R5} & {\footnotesize...As other reviewers have said, it's very easy to be here without a car - {\color{blue} balboa park is just across the road} and the airport is a short taxi ride away.} \\
\end{tabular}
\caption{Example showing a consensus or common opinion between 5 reviews for a hotel on TripAdvisor.com, taken from the SPACE corpus \cite{qt}}
\label{tab:introexample}
\end{table}

This lack of data for opinion summarization has motivated many abstractive summarization methods based on auto-encoders 
\cite{meansum,brazinskas-copycat,treestruct}, 
and these do not use any supervision from gold human summaries. A few recent
approaches propose self-training of encoder-decoder
models on synthetic summary examples. These examples are created by randomly sampling one of the 
input reviews and treating it as a pseudo-summary and treating other topically-related reviews as the source texts.\cite{amplayo-denoising,fewsum,amplayo-etal-2021-aspect,elsahar-etal-2021-self,brazinskas-etal-2022-efficient}
While such pseudo or silver-summaries are able to provide pretraining signals, their
objective is one of missing review prediction rather than aggregation of multiple texts. 
Their pseudo-summaries are also entire reviews which might contain other non-summary worthy content.

In this paper, we present a new self-training method which leverages
textual entailment signals to produce silver summaries of \emph{high quality and combining information
across multiple reviews}. Intuitively, the method aims to identify the consensus or most agreed
upon opinions in the source set. In the example in Table 1, if many reviews mention that the ``location is right across balboa park'', we would consider it a highly agreed upon opinion, and as a summary-worthy one. We create silver summaries using a set of such opinions with the highest agreement. 
We generate such silver summaries on a large scale and show how to train encoder-decoder transformer models using this data.
We evaluate our model in both zero-shot or unsupervised setting as well as few-shot learning. 
 Our results show that our method produces huge gains in both cases, outperforming other approaches and achieving new state-of-the-art performance. 




\section{Related work}



Opinion summarization is a widely studied problem, where the role of sentiment,
and product aspects (such as `lens' and `focus' for a camera) are well documented.
This paper focuses on abstractive summarization for general purpose summaries (not 
aspect-based) and we provide an overview
of the approaches closest to our work. 

\vspace{2mm}
\noindent{\bf Unsupervised neural networks.} As in other areas of text generation, modern opinion summarization methods are 
also predominantly neural networks. Since large scale data for supervision of 
encoder-decoder models is largely absent in this domain, many prior methods 
focused on unsupervised approaches. Common techniques here include 
auto-encoders and associated generative approaches such as VAEs. 

In \citet{meansum}, summaries are generated from the mean of the embeddings of 
input reviews. A similarity loss encourages the generated summaries to be
close to the review embeddings, while an autoencoder is used to 
improve the review embeddings. The intuition that summaries should 
capture the consensus opinions continues in \citet{brazinskas-copycat}, this
time employing a VAE that can steer towards common information. \citet{treestruct}
also use VAEs but extend them to produce hierarchical summaries where 
some sentences convey general impressions, while other provide specific details
about user experience. 

Our work also presents an unsupervised method, but based on encoder-decoder
models also taking advantage of self-training which we discuss next. 

\vspace{2mm}
\noindent{\bf Self-training methods.} Some very recent solutions have sought to take advantage of recent 
large pretrained encoder-decoder models via self-training \cite{amplayo-denoising,fewsum,amplayo-etal-2021-aspect,elsahar-etal-2021-self,brazinskas-etal-2022-efficient}. 
The approach here is to create large number of pairs of 
source review sets, paired with a pseudo or silver summary as an 
approximate target. In all these methods,
one of the reviews from the source
set is taken as the pseudo summary, and other reviews or topically 
related reviews to the target is taken as the set of source reviews. 
This dataset is then used for further pretraining of encoder-decoder
transformer models to incorporate signals and language 
specific to review summarization. These models are usually better
than unsupervised models based on generative approaches. 

While allowing a favorable paradigm shift, and better performance, 
there are a few limitations of this type of self-training. As pointed out
by \citet{fewsum}, reviews are considerably diverse from one another. So 
an objective that generates a review from other reviews will need to also
predict content not present on the source side, a major difference from 
actual summaries of reviews. Such pseudo-summaries will also contain 
a lot of first person language which again are less desirable in a summary to users. 

In this work, we present a novel method of pretraining. We also create 
silver-summaries on a large scale. However, our summaries actually contain
propositions from multiple input summaries and in particular those which 
are reflective of the consensus among the review authors. These summaries
are more powerful signals and move the training task away from review generation. 

\vspace{2mm}
\noindent {\bf Few-shot learning.} With increased use of encoder-decoder models, methods have also been 
proposed to efficiently augment the training with a small number of 
human-generated summaries (50 to 100). \citet{pass} train 
transformer models on a small number of examples and during inference, 
generate multiple summaries which are later ranked according to coherence 
to arrive at a final one. Other approaches focus on 
an additional plug-in network that can predict desired properties of
summaries based on a few labelled examples \cite{fewsum} that can 
augment training signals. \citet{brazinskas-etal-2022-efficient} 
introduce the use of a few additional parameters in the form of adaptors
and only these are finetuned instead of the full network, making the training
efficient and robust for few-shot learning. 

We also demonstrate our self-trained model in few-shot settings.

\vspace{2mm}
\noindent{\bf Consensus as a goal for summarization.} 
When the summarization problem contains multiple input texts, 
intuitively the frequently held or common information across
them is one important signal for summary worthy content. 
Multi-document news summarization has exploited frequency
from early times \cite{sumbasic,radev2004} to most 
recent ones \cite{ernst-etal-2022-proposition}. Recent work has also used consensus
as a goal for summarizing scientific publications
around health topics \cite{nutribullets}, and identify 
agreement and discrepancies in Wikipedia document clusters \cite{sentnli}.

Intuitively, review
summarization also expects to capture the voice of the majority of users
as one of its aims. For example, if a majority of users complain about the battery of an item, we 
would expect a summary to mention that. Instructions to annotators in multiple
annotation efforts for opinion summarization explicitly ask annotators to 
capture what is common and popular \cite{fewsum,qt}. The idea of consensus
is also present in the objective of many recent models for opinion
summarization \cite{meansum,brazinskas-copycat,qt}. In this work, our 
self-training approach explicitly tries to capture statements which 
are agreed upon by a majority of reviews.

\section{Textual entailment to identify consensus among review users}
\label{sec:silverdata}

We propose a novel approach to create silver source-summary pairs for
abstractive opinion summarization. A central idea here is the use of 
textual entailment to find statements reflecting user consensus. We
first present our definition of the idea and describe the 
steps involved in silver data creation. 

\subsection{Defining review consensus}

We define consensus as the number of reviews that support a particular claim. For example, 60 (out of 100) reviews
might claim that the seafood dishes are great at a restaurant. Likewise 30 reviews
might say that the staff are friendly and polite. Our aim is to obtain those 
sentences with most user consensus automatically,
and use these to create our silver-standard data.

But note that the same claim may be expressed in different ways or granularity, and so 
their frequency in reviews cannot be easily computed. 
Eg. \emph{`This hotel is in the heart of Times Square'} and 
\emph{`Hotel's location is slap bang in the middle of 
Times Square.'} both express the same claim, and 
\emph{`The fish is tasty'} and \emph{`The salmon is delicious'}, both support the
claim that \emph{`The seafood is great.'}. 
Our idea is to capture this variability using natural language entailment. 

At a high level, our approach identifies potential claims in the form of 
\emph{propositions} from a large collection of texts,
uses textual entailment to find out how often the collection supports the proposition,
and computes a score for the support. 
	
Now we explain how we obtain these statements and their scores automatically.


\subsection{Extracting propositions}

For texts, even when they are sentence-level units, it is hard to 
reason about them precisely. Many 
review sentences in addition tend to be rather long. For example, 
\emph{``I love eating in Likya, the Chefs are so passionate and professional 
about the food they cook and the staffs are well trained, they treat me 
very well like a customer.''} contain a bunch of different claims. 
It is difficult to find support for such complex sentences since 
the same information is unlikely to be present in other users' reviews.

Instead, we split review sentences into \emph{propositions} and use these as 
our key units. We define a proposition as a `single claim or fact' about the item and 
extract these as snippets from the original review texts. In fact, recent work 
on supervised news summarization also uses the extraction and clustering of 
proposition units to find frequent subtopics, and then fusing the information in the
biggest clusters into a summary \cite{dagan-proposition}.

In this work, we use simple rules to split review sentences
into propositions. We split sentences at conjunctions, period, and comma subject
to a minimum clause length of four. Our algorithm processes sentences from left to 
right to find a delimiter. If the proposed span will create a clause less than the minimum
length, we do not split and attach the span to the proposition on the left. 
Note that these propositions
are a linear segmentation of the input sentence,
and their concatenation yields the original 
sentence. Intuitively, this process primarily performs
syntactic simplication, without changing the total content
that is expressed. 

The resulting propositions for different sentences
in our data is shown in Table \ref{tab:proposition_splitting}. Note that there are some propositions which
end up ungrammatical, and our length constraints do not always separate out
all the aspects (as in the third example in Table \ref{tab:proposition_splitting}).
But overall this simple method works well for review sentences where 
syntactic embedding is less complex than in genres such as news, and we can scale to 
large collections efficiently. 

\begin{table*}[ht!]
    \centering
    \begin{tabular}{|p{6cm}|p{8cm}|} \hline
    {\footnotesize \bf Review sentence} & {\footnotesize \bf Extracted propositions}\\ \hline

{\footnotesize There was loads of cupboard space and a fantastic easy to use safe.} & {\footnotesize There was loads of cupboard space and$_1$ a fantastic easy to use safe.$_2$} \\ \hline
{\footnotesize Metro station (llcuna, line 4) is 5 minute walk away, beach is a 10 minute walk away.} & {\footnotesize Metro station (llcuna, line 4) is 5 minute walk away,$_1$ beach is a 10 minute walk away.$_2$}\\\hline
{\footnotesize The room was very nice and clean, quiet location, staff were helpful, easy access to the centre of town by metro, bakeries and a supermarket nearby.} &  {\footnotesize The room was very nice and clean, quiet location, staff were helpful,$_1$ – easy access to the centre of town by metro, bakeries and a supermarket nearby.$_2$} \\\hline
    \end{tabular}
    \caption{Example propositions split from source sentences. The propositions on the right are numbered according to their position in the sentence.}
    \label{tab:proposition_splitting}
\end{table*}




We extract propositions from all the reviews for an item. 
Suppose there are $N$ reviews for
an item which result in $M$ propositions where $M \gg N$. 

\subsection{Scoring consensus}

Our aim is to find the number of supporting reviews for each of the $M$ propositions. We compute
this number using natural language entailment. Specifically, consider review $R_i$ and proposition $m_j$ belonging to the
same item. Let us represent a textual entailment relation as $P \rightarrow H$, where $P$ is a
premise and $H$ is a hypothesis. In our case, if
$R_i \rightarrow m_j$, then we consider that $R_i$ \textit{supports} $m_j$. The
final score for proposition $m_j$, $S(m_j) = \sum_{1 \le i \le N}E(R_i, m_j)$ where $E(R_i, m_j)=1$
if $R_i \rightarrow m_j$ else 0. 

We obtain $E(R_i, m_j)$ using the predictions of an entailment classifier
which treats the $R_i$ as the premise and $m_j$ as the hypothesis.
If the most likely label from the classifier is 
`entailment', then $E(R_i, m_j)=1$ and 0 if other labels had the highest probability. 

In this work, we use a cross attention model, BERT-large \cite{devlin-etal-2019-bert} to obtain these predictions. The input to the model concatenates the premise and hypothesis with a separator symbol and the CLS token's embedding is sent through a linear layer to predict three classes: entailment, contradiction and neutral. We trained this model on the MNLI corpus \cite{mnli} reaching a development accuracy of 84\%. Note that
the training data for the entailment model does not contain any examples from  the review domain. But we found that predictions are rather reasonable and even better when a higher threshold is applied on the probability of the entailment label. 

Note that this score computation for all propositions requires an entailment prediction between all pairs of $(R_i, m_j)$. Even though the computation is
done only within each item, there are still a quadratic number of pairs per item. 

So we implement the full computation of silver summaries in a Apache Beam\footnote{\url{https://beam.apache.org/}} pipeline which allows to create parallel data-processing pipelines. Our typical pipelines do inference billions of times by the entailment models. 



In Table \ref{tab:entailmentset}, we show some of the entailment predictions from our models. We take a proposition and sample random reviews from the set of reviews which entail that proposition. Our model does not explicitly do any sentiment classification, we have picked a positive and negative proposition for demonstrating how precise and clear entailment based support prediction tends to be. 

\begin{table*}[ht!]
    \centering
    \begin{tabular}{|p{15cm}|} \hline
    {\footnotesize \bf Proposition: ``the property has a lot of character''} \\
    {\footnotesize \bf Supporting reviews:}\\
    {\footnotesize R1. ...Though i understand the previous posters point that the park manor has charm, I'd say that the actual ``charm'' happens in all the wrong places. That there's a nice and funky lobby with some amazing artistic featurettes and a cute patio with a coy boy, or the spacious rooms with a hodgepodge of furniture and beautiful molding on the walls that seems to go nowhere - yes, charming.}\\
    {\footnotesize R2. ...but the views higher would have been spectacular. A quirky place which people will love or hate...} \\
    {\footnotesize R3. ...this hotel is beautiful! It 's so elegantly decorted but in an antique way. The ceiling in the lobby... a huge king bed, sofa, armoire, vanity desk, kitchen - stove, refridgerator and the necessary kitchenware. I loved all the antique furniture, so nice to look at and change from standard hotel decor...} \\
    {\footnotesize R4. ...I would highly recommend this hotel to anyone who is looking for accommodations with more character than you 'll find at the big chain hotels. A marriott looks like a marriott whether you're in singapore or st. Louis. Why not try the local flavor?...}\\
    {\footnotesize R5. ...This hotel is old and dated. The furnishings are very old and the whole hotel needs refurbishing . there are gas stoves in the rooms...}\\ \hline
    {\footnotesize \bf Proposition: ``obvious neglect to fixtures and fittings.''} \\
    {\footnotesize \bf Supporting reviews:} \\
    {\footnotesize R1. ...i leant on the bannister at one point and almost fell down three floors...the window would not close... the electricity in our room kept cutting out if we had more than one item on...}\\
    {\footnotesize R2. ...my friends also got two leaks in their room... the carpets were old and they were obviously never hoovered in years...i saying they should knock the building down and do the whole thing up... }\\
    {\footnotesize R3. ...there were loose electric wires hanging from the ceilings-which i tripped over constantly...the locks on the doors were poor... }\\
    {\footnotesize R4. ...there are no elevators and the stairs are falling apart- literally!...broken window which was taped up with parcel tape and cardboard... broken heaters...wardrobe with door falling off... }\\
    {\footnotesize R5. ...could not charge phones because outlets did not work...cable tv was finnecky...internet was one computer on the second floor and did not work most of the time...broken fixtures and missing electrical covers...building seemed to be crumbling and it leaked in the foyer when it rained...}\\ \hline
    %
    \end{tabular}
    \caption{Two example propositions (from two hotels in our dataset) with 5 reviews which entail them. The reviews were randomly selected from the full list of reviews which entail each proposition.}
    \label{tab:entailmentset}
\end{table*}

\subsection{Silver summaries}

We order the propositions in decreasing order of their scores $S(m_i)$,
and take the top $n$ as the silver summary sentences. We trim the silver summary up to a certain 
summary length expressed in tokens. Additionally, we employ a MMR \cite{mmr} style redundancy removal
technique to keep diverse content in the summary. We implement this control using a simple method of content word overlap.\footnote{We also explored
entailment based diversity measures, but we found that simple content word overlap kept the maximum diversity in aspects commented on within the summaries.} Suppose $S$ is the set of propositions selected in the summary so far. The next proposition chosen 
is the highest scoring proposition $p_k$ where $overlap(p_k, s_i) < 2$, 
$\forall i, 1 \leq i \leq \lvert S \rvert$. $overlap$ is computed as the number of content words in common between the two propositions based on the stopword list within NLTK \cite{nltk}. 

The top propositions for two hotel items  from our dataset is shown in Table \ref{tab:top_propositions}. Note that these
are snippets from actual reviews for that item or product. 

\begin{table*}
\centering
\begin{tabular}{p{6.5cm}|p{8.5cm}}\hline
{\footnotesize Hotel with 106 reviews} & {\footnotesize Hotel with 61 reviews} \\ \hline
{\footnotesize 1. very comfortable (a big deal for me). (58\%)} & {\footnotesize 1. well equipped with good privacy setting. (82\%)}\\
{\footnotesize 2. well maintained, clean, comfortable suites, (57\%)} & {\footnotesize 2. the family-owned vacation spot is very family oriented. (68\%)}\\
{\footnotesize 3. the rooms were very comfortable, and (55\%)} & {\footnotesize 3. this resort is a comfortable family retreat providing a great getaway. (60\%)}\\
{\footnotesize 4. they have a place to eat but (52\%)} & {\footnotesize 4. a very family friendly place to stay. (60\%)}\\
{\footnotesize 5. the size of the room is nice, (51\%)} &{\footnotesize 5. our unit was very clean, comfortable.. (55\%)}\\
{\footnotesize 6. that was a great rate for a suite. (50\%)} & {\footnotesize 6. units have had great proximity to a pool and (54\%)}\\
{\footnotesize 7. still professional; the room was clean and (50\%)} & {\footnotesize }\\ \hline
\end{tabular}
\caption{The top propositions for two hotels in our dataset. We take the top 10 propositions and show only the ones kept after redundancy filtering. The percentage of total reviews which entail each proposition is shown within braces.}
\label{tab:top_propositions}
\end{table*}

This final set of summary propositions, $S$, chosen for a given summary length,
are then concatenated in the chosen order to create the silver summary. When the
propositions are not full sentences, we polish them for capitalization and 
punctuation to match full summaries. Note that no special facility is present for ordering these sentences by coherence. In many cases, the list of top propositions is a very reasonable summary, and in this first work, we have not carried out further processing for coherence.


\subsection{Source texts}
\label{sec:reviewsampling}
The silver summaries from the previous step are 
composed of extracted text spans from the source reviews. A system trained to 
produce such sequences from the full set of input
reviews will predominantly copy from the input
texts. So we make changes to the set of source reviews to turn the data into one suitable for abstractive summarization. 

\begin{figure}
    \centering
    \includegraphics[width=0.45\textwidth]{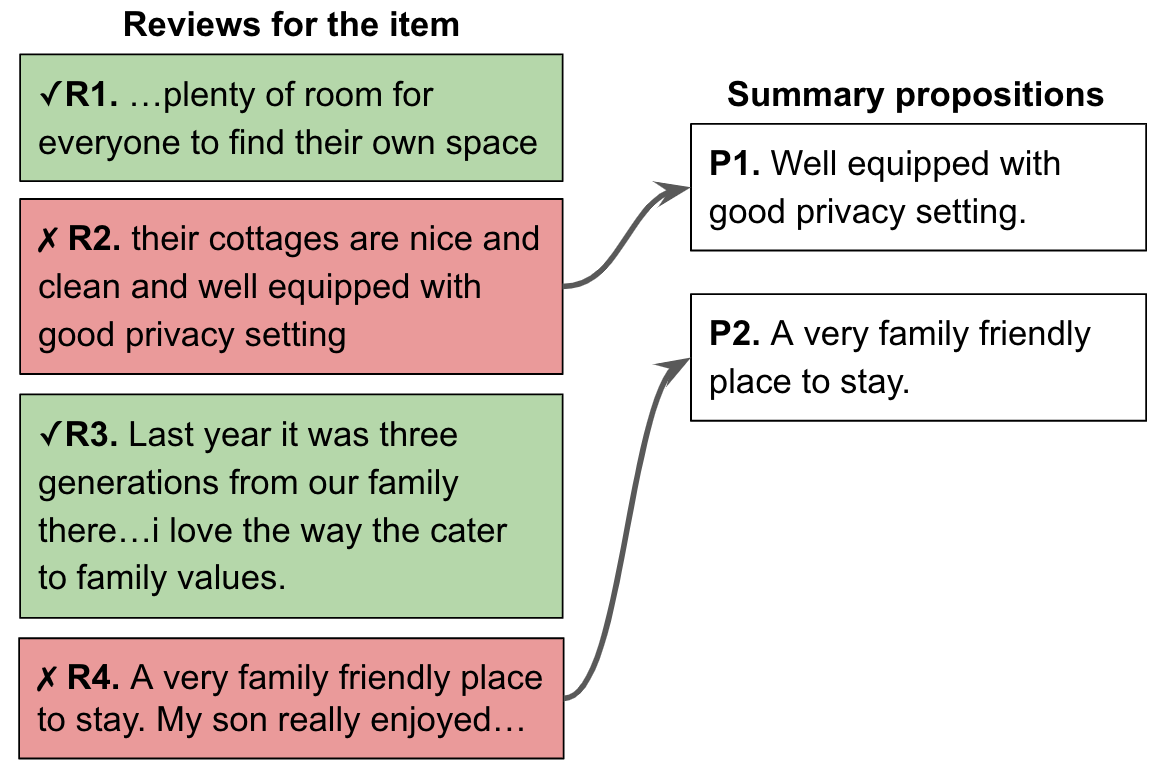}
    \caption{Example which demonstrates how 
    reviews are removed from the summarization 
    input side if they were the original source from which a proposition was extracted. Here, P1 was extracted from R2 and P2 from R4. R2 and R4 will be removed entirely from the summarization input. But note that the summary content is present in other reviews which entail P1 and P2.}
    \label{fig:source_masking}
\end{figure}

Let $N$ be the total set of input reviews. 
For each proposition $p_i$ in the summary, we remove the review $R_j$,
where $p_i$ came from, i.e. $p_i$ is a span in $R_{j}$. This 
deletion discourages the verbatim copying of text spans from the source, 
and encourages systems to perform abstraction. 
The final input reviews on the source
side is a reduced set $N'$, $|N'|<|N|$. 
Note that sentences (propositions) in the silver standard are 
supported by many other reviews, albeit in different surface forms,
so the signals to produce the silver summary are still present in $N'$. An illustration of 
input review selections is shown in Figure \ref{fig:source_masking}. This 
way of creating source-summary pairs resembles one of the 
powerful pretraining objectives for abstractive summarization
known as Gap Sentences Generation, introduced
by the Pegasus \cite{pegasus} model.

In practice, the number of source reviews that can be processed as input in most standard sequence to sequence models is fewer than the hundreds present in $N'$.
So we sample a smaller set $N''$, size $k$, 
of reviews to fit the sequence length of the encoder. 
We could further aid the training by adapting $N''$
to be most useful for generating the target
sentences.  We can create this
sample of size $k$ in three ways. 

{\sc Uniform.} In the simplest case, we can 
sample $k$ source reviews uniformly at random.

The remaining two methods focus on those reviews which entail one of the silver-summary propositions.

{\sc Equal.}  We sample $k/|S|$ reviews from the set of reviews entailing
each proposition in the summary. The intuition here
is that the summarization source contains an equal number of reviews supporting each (silver) summary proposition. 

{\sc Proportional.} We sample $l$ reviews from the entailment set of 
each summary proposition, where $l$ is proportional to the size of 
the entailment set. For example, if `seafood is great' is a summary proposition with 40\% entailment support, then 40\% of the summarization input are reviews on the topic of great seafood, although the review containing the verbatim proposition is filtered out.


In the next sections, we describe how we use this data to train 
abstractive summarization systems. 






\section{Datasets}
\label{sec:datasets}


We use two sources of data in our experiments. 
The first is an unlabelled review corpus (no gold or human summaries are available for the items). This dataset is used to create silver-standard
summaries for self-training. The second source is an evaluation dataset containing a much smaller set of items (not seen during training) and here for each item, the set of source reviews are paired with one or more human summaries of those reviews. 



In our experiments we use the SPACE
corpus collected by \cite{qt}. It comprises of reviews for hotels from the
TripAdvisor website. 


\vspace{2mm}
\noindent{\bf SPACE-unlabelled.} This is 
a collection of 1.1 million reviews for 11,000 hotels. These reviews are not paired
with human summaries. We use this set for silver data creation and further for training. 


\vspace{2mm}
\noindent{\bf SPACE-eval.} contains human generated summaries for a smaller set of 50 hotels. For each of these hotels, 100 input reviews are paired with 3 gold-standard
human-written abstractive summaries.  The human-summaries were 
created via a two-step process where annotators first
selected key sentences from the input reviews, and then wrote a summary
based on the sentence collection. The dataset contains 3 general summaries for each hotel, as well as aspect based 
such as for food and cleanliness. We only
use the general summaries for each input. 
These 50 hotels are divided
into 25 for development and 25 for test sets.

This evaluation dataset ideally suits our
task since the input contains 100 reviews on which 
one could ask for common opinions and claims. Most other evaluation sets \cite{brazinskas-copycat,fewsum} contain about 8 randomly sampled reviews which may often not have much in common.

\section{Models}

We build our abstractive systems using pretrained encoder-decoder models based on T5's \cite{t5} framework. 
These models encode the input reviews as a sequence and autoregressively generate
the output summary words as a sequence. 

In multi-document summarization, especially opinions, source reviews could easily span
hundreds of reviews. Standard self-attention layers found in current transformer models have a polynomial scale relationship to input length, making it impossible to encode and attend to several reviews at once. Many summarization systems avoid this issue by including a content selection component as a first step of a pipeline.
Recent work has shown that sparse transformers are able to overcome this issue, simplifying models and many times outperforming 
pipeline based alternatives. For this reason, we have also built models on top of LongT5 \cite{longt5}, which implements sparse attention by combining
local attention with transient global attention, allowing tokens to attend locally and globally via transient global nodes.

In this work, we employ
LongT5 models (of different sizes: Large (770M), XL (3B)) with a limit of 8,192 sentence pieces.
We use the public pretrained checkpoint. \footnote{
\url{https://github.com/google-research/longt5}}

\section{Experiments}

In this section, we explain how we trained our abstractive summarization models.

\subsection{Silver Data}



We create our silver data using the unlabelled review corpus introduced 
in Section \ref{sec:datasets}. We called this silver dataset as {\bf SPACE-OpineSum}.

To create this set, we followed the procedure outlined in \ref{sec:silverdata}. We used SPACE items with a 
minimum of 50 reviews (since very few reviews may not have a lot in common to extract out). This set contains about 
4,729 items. Our beam pipelines computed a total of around 1.3B
entailment predictions on review-proposition pairs from these items. The resulting silver
data contains the same number of items, but now each item is paired with a silver summary.


\subsection{Self-training}

We explore the usefulness of our self-training in two setups: unsupervised and
few-shot learning abstractive summarization. For the unsupervised case, 
we train our models on the silver-data only. For few-shot learning, we 
use a small number of annotated input-summary pairs ($<$100) for finetuning
our self-supervised systems.

\subsubsection{Unsupervised training}

Given the silver-data, we trained LongT5-Large (770M parameters) and LongT5-(Large, XL) \cite{longt5} models on the sequence-to-sequence
task of generating the highest consensus opinions (i.e. most entailed) given 
a concatenated sequence of the input reviews. These models do not use 
any gold-annotated examples for training.

We compare these systems with prior unsupervised work in the SPACE-eval 
dataset introduced in Section \ref{sec:datasets}. We select
the best checkpoint based the ROUGE performance on the validation set.

\subsection{Few-shot Learning}
\label{sec:fewshot}

Few-shot learning was implemented by finetuning our self-trained models 
on a few human annotated source-review and summary pairs. 
To facilitate this setup, we divide
the development examples in SPACE-eval (25 total) into a training set with 15 items and a validation set with 10 items. The test set remains unchanged.
We use this training set for few-shot learning and the best checkpoint was selected based on ROUGE scores on the 
validation set.
These models trained better with a rather reduced learning rate, $1/5th$ of the standard $1e-4$. 

We will compare these models with baselines which do not use self-training
with silver summaries. Rather
these latter models are warm started 
from the public pretrained checkpoints and similarly trained on the train split we created above.  


\section{Results}

First we present which settings were most useful
for self-training before describing summarization performance. 

One aspect is the relationship between input source reviews and the silver summary. We trained all our models until
validation performance plateaus. In this case, ROUGE was computed on the held-out validation silver data set.

In Section \ref{sec:reviewsampling}, we present three ways of sampling the set of 
source reviews to consider as input: equal, uniform, and proportional. We found that our model performance was similar across these settings. Since our output propositions are only a list, perhaps a model can learn the relationship as long as there are frequency signals in the input, but that frequency does not need to be proportional to the frequency seen in the full set of input reviews. 



We also compared how many reviews, size $k$, should be present on the input side. While there were no strong patterns as for sampling methods, typically more reviews, eg. 160 performed better most of the time. 
Next we compare how well the models perform in the unsupervised summarization 
setting. Here we train our models on the silver data and evaluate on the test set of SPACE-eval. 
 Table \ref{tab:results_unsupervised} presents 
the ROUGE scores. We compare with previous Lexrank \cite{erkan2004lexrank} results as well as the current best system ACESUM by \cite{amplayo-etal-2021-aspect}. 

We see that {\sc OpineSum} systems obtain very good performance.
Sometimes we do not outperform the best state of art system since these
systems are sophisticated and
tend to employ a variety of techniques (such as aspect extraction)
while our model is only driven 
by self-training. We would expect that the addition of other modules would improve upon our system. 

\begin{table}[t]
    \centering
    \begin{tabular}{r|ccc}
    {\bf Model}    & {\bf R1} & {\bf R2} & {\bf RL}\\
    \hline
    \multicolumn{4}{c}{Previous systems} \\
    \hline
     Lexrank        & 36.86  &  8.81 & 22.96 \\
     Acesum         & 42.64  & 14.50 & 25.20\\\hline
     \multicolumn{4}{c}{ {\sc OpineSum} systems} \\ \hline
     LongT5 Large  & {\bf 45.84} & \bf{16.30}  & {\bf  29.18} \\
     LongT5 XL         &  43.41          &    13.82       &   23.84     \\ \hline
    \end{tabular}
    \caption{Results for the unsupervised setting. The {\sc OpineSum} systems use self-training only and {\em no gold summaries}.}
    \label{tab:results_unsupervised}
\end{table}

Table \ref{tab:results_fewshot} presents results in the few-shot learning setup. 
There are no prior system results for 
this few-shot setup on the SPACE data.
Nevertheless, the 
T5 models trained without silver-data 
are a strong ablation to compare against our few-shot trained models with {\sc OpineSum} warm start. 

Here, we 
see that the baseline T5 examples are already rather strong and outperform earlier
unsupervised systems. 
In particular, LongT5 is pre-trained with a summarization-relevant objective: the gap sentence prediction task. That is a probably cause for its high performance on this task. Even with this high baseline, we find that our simple self-training still leads to further significant improvements.

We show an example output of our system compared with gold standards and prior system in Table \ref{tab:example_outputs}. One noteworthy difference is between our unsupervised and fewshot systems. The unsupervised system produces shorter summaries and at time disfluencies due to being trained on smoothed propositions. Fewshot learning improves along these dimensions being the summary much closer to the gold standards. Also note that ACESUM summaries contain many phrasal repetitions while that is absent in our outputs. 


\begin{table*}[h!]
    \centering
    \begin{tabular}{|p{15cm}|} \hline
    {\footnotesize \bf Gold standard summaries} \\
 {\footnotesize G1. This hotel was very nice and within walking distance of the Vatican, Colosseum, Forum, ST Peters, etc. Staff were helpful in every way, and the attention to each request and question was efficient and treated with courtesy. The air-conditioned rooms were very nice, clean, and comfortable, with immaculate bathrooms to boot. Breakfast, which is included, was pretty good for a continental buffet.} \\ 
 \\
{\footnotesize G2. Staff received mixed reviews, but were overall considered friendly, attentive, and helpful. The hotel, rooms, and bathrooms were very clean, with daily maid service and linen change. The room was beautiful and airy. The Breakfast was great and varied. The location is excellent, away from the hordes of tourists. It's just a short walk over Ponte Umberto to Piazza Navona, or across Ponte Cavour to reach the popular shopping areas. The building is nice. The restaurant was first rate. However, some thought that the hotel is pricey for the quality of the room.}\\ 
\\
{\footnotesize G3. The staff was extremely courteous and helpful. The wooden floors were all cleaned and maintained; as well as everything else in the hotel. The rooms were beautiful and large, and the bathroom was immaculate. There was a good, buffet style breakfast with particularly enjoyable cold meats, and with anything else desired. The hotel is located close enough to the Vatican, Colosseum, the Forum, and St. Peters- overall a great location.} \\ \hline
\\
{\footnotesize {\bf ACESUM} \cite{amplayo-etal-2021-aspect}}\\
{\footnotesize The staff were very friendly and helpful. the room was clean and clean. it was a great place to stay. if you want to stay in cicerone, it is a good place to get to the shopping area. there are many restaurants, restaurants and restaurants. but the staff are very friendly, friendly and friendly. they are a very nice hotel, a nice place to eat, and a lot of good food, as well as a small restaurant, the breakfast was very good, but a bit of.} \\ 
\\\hline
{\footnotesize \bf {\sc OpineSum}-unsupervised} \\
{\footnotesize  The hotel is located within walking distance of the Vatican. The rooms were clean and comfortable. The Cicerone is a nice hotel. As far as the hotel goes. The reception area is nice but the rooms. The breakfast buffet was fine. The room was a good size.} \\
\\\hline
{\footnotesize \bf {\sc OpineSum}-fewshot}\\
{\footnotesize The staff was friendly and helpful. The rooms and hotel itself is modern, extremely clean! The rooms are a good size, with comfy beds, a breadth of amenities such as a great shower and a comfortable bed. The breakfast buffet is average, but very good, with lots of variety. The location is very central. The hotel is within walking distance of the Vatican and Piazza Navona. The Cicerone is a beautiful hotel, but the hallways need refurbishing.}\\ \hline
 \end{tabular}
 \caption{Example summaries for one item in our dataset. We show the 3 gold standard summaries available on the evaluation set along with previous best system (ACESUM) and our unsupervised and few-shot self-trained systems.}
 \label{tab:example_outputs}
 \end{table*}
 
\begin{table}
    \centering
    \begin{tabular}{rc|ccc}
    {\bf Model}      & {\bf {\footnotesize Checkpoint}} & {\bf R1} & {\bf R2} & {\bf RL}\\ 
    \hline
     {\footnotesize LongT5 L}    &   {\footnotesize Vanilla}    &  45.51   & 13.03 & 29.28\\ 
     {\footnotesize LongT5 L}    &   {\footnotesize \sc{OpineSum}}   &  {\bf 47.19}  & {\bf 14.60}   & {\bf 30.13} \\
     \hline
    \end{tabular}
    \caption{Results for the few-shot learning setting. All the models
    were finetuned on a small set of 15 training examples 
    described in Section \ref{sec:fewshot}. `Vanilla'
    systems are warm started from public checkpoints and do not see self-training data.}
    \label{tab:results_fewshot}
\end{table}

\section{Conclusion}

We have presented a simple self-training approach which leads to sizeable gains on both unsupervised and few-shot abstractive opinion summarization. 

Our work is one of the first to demonstrate how an intuitive idea of consensus can be incorporated during self-training. It opens up a number of challenges and new problems for future work. In particular, while our silver data contains the provenance for each top proposition---meaning the set of reviews which support each the proposition---this information is only minimally used at the moment. Future work could explore how models could be trained using the entailment weights (scores) of each proposition and the exact links to entailing reviews to yield more performance improvements and faithful generation.

We also hope that such self-training models could serve as good checkpoints for other tasks in the opinion domain such as review helpfulness prediction or product popularity and ratings. We hope to explore such directions in future work.

\bibliographystyle{acl_natbib}
\bibliography{custom}

\end{document}